\documentclass{article}
\usepackage{arxiv}

\title{Conversational Medical AI: Ready for Practice}

\date{}

\author{
    \begin{center}
        \parbox{0.8\textwidth}{
            \centering
            \large
            \href{https://orcid.org/0000-0002-1073-3190}{\includegraphics[scale=0.06]{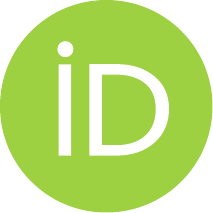}}\hspace{1mm}Antoine Lizée\footnotemark[1]\customfootnote,
            Pierre-Auguste Beaucoté\footnotemark[1],
            James Whitbeck\footnotemark[1],\\
            Marion Doumeingts\footnotemark[1],
            Anaël Beaugnon\footnotemark[1],
            \href{https://orcid.org/0000-0002-3170-3978}{\includegraphics[scale=0.06]{helpers/orcid.pdf}}\hspace{1mm}Isabelle Feldhaus\footnotemark[2]\\
        }
    \end{center}
}

\hypersetup{
pdftitle={Conversational Medical AI: Ready for Practice},
pdfsubject={q-bio.NC, q-bio.QM},
pdfauthor={Antoine Lizée, Pierre-Auguste Beaucote, James Whitbeck, Marion Doumeingts, Anaël Beaugnon, Isabelle Feldhaus},
pdfkeywords={Conversational AI, Medical AI, Patient Experience, Healthcare Communication, Large Language Models, Clinical Safety, Digital Health, Medical Chat, Patient Satisfaction, Healthcare Access, Physician Oversight, Real-world Evaluation, AI Implementation},
}

\begin{document}

\maketitle

\renewcommand{\thefootnote}{\customfootnote}
\footnotetext[4]{Corresponding Author: \href{mailto:antoine.lizee@alan.eu}{antoine.lizee@alan.eu}}
\renewcommand{\thefootnote}{\fnsymbol{footnote}} 
\footnotetext[1]{Alan, France}
\footnotetext[2]{Belle Labs, France}

\begin{abstract}

The shortage of doctors is creating a critical squeeze in access to medical expertise. While conversational Artificial Intelligence (AI) holds promise in addressing this problem, its safe deployment in patient-facing roles remains largely unexplored in real-world medical settings. We present the first large-scale evaluation of a physician-supervised LLM-based conversational agent in a real-world medical setting.

Our agent, \mo, was integrated into an existing medical advice chat service. Over a three-week period, we conducted a randomized controlled experiment with 926 cases to evaluate patient experience and satisfaction. Among these, \mo\ handled 298 complete patient interactions, for which we report physician-assessed measures of safety and medical accuracy.

Patients reported higher clarity of information (3.73 vs 3.62 out of 4, p < 0.05) and overall satisfaction (4.58 vs 4.42 out of 5, p < 0.05) with AI-assisted conversations compared to standard care, while showing equivalent levels of trust and perceived empathy. The high opt-in rate (81\% among respondents) exceeded previous benchmarks for AI acceptance in healthcare. Physician oversight ensured safety, with 95\% of conversations rated as “good” or “excellent” by general practitioners experienced in operating a medical advice chat service.

Our findings demonstrate that carefully implemented AI medical assistants can enhance patient experience while maintaining safety standards through physician supervision. This work provides empirical evidence for the feasibility of AI deployment in healthcare communication and insights into the requirements for successful integration into existing healthcare services.

\end{abstract}


\section{Introduction}
\label{sec:introduction}
Globally, persistent shortages and inequitable distribution of the health workforce contribute to decreased access to health services and poorer quality of care. Projections indicate a shortage of 10 million health workers worldwide by 2030 \cite{Boniol2022}. Countries across Europe are facing shortages in primary care physicians, aggravated by aging populations and increased chronic disease burden \cite{Russo2023}. Regional disparities are particularly pronounced, with urban areas generally having higher physician densities than rural regions \cite{Winkelmann2020,Pl2021}. Studies report deteriorating access to care, especially in these underserved areas, leading to increased workloads and burnout among practitioners \cite{Dumesnil2024}. Physician burnout is associated with reduced engagement and lower quality of care \cite{Zhou2020}. The limited availability of primary care services not only restricts access to preventive and routine care, but also creates additional strain on emergency services, ultimately degrading the overall quality of care \cite{Russo2023}. 

While the successful deployment of machine learning and Artificial Intelligence (AI) in healthcare settings is not new, these technologies are typically not directly engaged in patient care and communications. Their functions have been largely reserved for expert use in signal processing, predictive analytics, medical image analysis, and medical devices innovations \cite{Rubinger2023,Susanto2023,Lyell2021}. 

Recent advances in general-purpose large language models (LLMs) and generative AI have opened new opportunities for healthcare applications, particularly through conversational AI agents optimized for medical use \cite{tu2024conversationaldiagnosticai}. Such agents can serve a number of critical roles fundamental to a patient’s care, health literacy, coordination, and management. By directly answering patients’ medical questions more readily, collecting relevant diagnostic information, and facilitating patient-provider communication, they could help address the growing challenges in access and quality of care. This potential has prompted active research into the safety, accuracy, and effectiveness of conversational AI agents in healthcare settings. 

Retrospective and modeling analyses show that AI agents perform increasingly well on metrics evaluating diagnostic accuracy, answers to patient-directed medical questions, knowledge recall, and medical reasoning \cite{tu2024conversationaldiagnosticai,Zeltzer2023,singhal2023expertlevelmedicalquestionanswering,Singhal2023}. In Tu et al. (2024), AMIE (Articulate Medical Intelligence Explorer), an LLM-based AI system optimized for clinical history-taking and diagnostic dialogue, demonstrated greater diagnostic accuracy and superior performance compared to physicians in simulated consultations with patient actors \cite{tu2024conversationaldiagnosticai}. Evaluating the safety and performance of patient-facing conversational AI agents in a real-world setting is among the next steps forward.

Alan, a health and insurance company operating in France, Belgium, Spain, and Canada, has offered a medical chat advice service to its members since 2020. Using the Alan mobile app, any Alan member can ask a question directly to an on-call physician through the privacy-compliant chat. In 2024, Alan introduced \mo, an LLM-based conversational agent, to this medical advice chat service staffed by its general practitioners. 

In this study, we present our findings from this experiment in introducing conversational AI into medical practice.

Our primary contributions are:
\begin{itemize}
    \item We introduced \mo, a patient-facing medical agent designed as an AI system. To this end, we developed a comprehensive evaluation framework combining clinical knowledge and reasoning assessment, real-world conversation analysis, and automated testing through simulated patient interactions.
    \item We integrated \mo\ into a pre-existing medical advice chat service, with a focus on ethical design for patients, physician oversight, and quality assurance.
    \item We ran a randomized controlled experiment, collecting data over 3 weeks to compare patient satisfaction and experience between conversations when \mo\ was proposed and a control group of patients that interacted solely with human physicians. The experiment highlighted that overall satisfaction and perceived clarity were higher in conversations with \mo, while trust in the received information and perceptions of empathy were similar between the two groups. We also show that patient engagement is higher in conversations with \mo, evidenced by shorter response times from patients.
    \item We evaluated safety and medical accuracy through physician reviews. 95\% of the conversations were assessed as “good” or “excellent”, while no conversation was considered as potentially dangerous overall.
    \item Finally, we discussed the implications of our findings for the broader adoption of AI in healthcare, focusing on patient empowerment, access to care, and the evolution of healthcare delivery models.
\end{itemize}

\section{\mo, an LLM-based medical conversational agent deployed in Alan’s medical chat}
\label{sec: mo, an LLM-based medical conversational agent deployed in Alan’s medical chat}

\subsection{Context}
\label{subsec:context}
Alan is a health and insurance company established in 2016 and headquartered in Paris, France. With operations across France, Belgium, Spain, and Canada, Alan provides health coverage for approximately 700,000 members as of October 2024. To accomplish its mission of making health simpler, transparent, and accessible for all of its members, the company designs, develops, and releases innovative digital products for the personalized use of its members. This capacity is built on Alan's dual expertise in technology (i.e., software engineering and research) and healthcare, allowing the company to build digital solutions that serve members’ health needs. 

In 2020, Alan introduced a medical advice chat service as a way to enhance its product and service offerings for its members. Using the Alan mobile app, members can directly contact a general practitioner or specialist physician to receive answers to their medical questions during extended hours (from 7 am to 12 am, seven days a week). The medical advice chat service is fully compliant with health privacy regulations in France and the European Union (EU), and uses end-to-end encryption for the messages between members and physicians.

Between January 1 and October 1, 2024, Alan's medical advice chat service facilitated over 58,000 conversations between members and health professionals. These conversations were split between general practitioners (62\%) and other healthcare professionals specializing in physiotherapy, nutrition, gynecology, pediatrics, dermatology and sexual health. At the beginning of the study, general practitioners (GPs) had been operating the service for an average of 2.8 years (range: 0.8 - 4.0). Towards supporting the doctors operating the service, Alan introduced an LLM-based conversational agent into its medical advice chat service over the summer of 2024.

\subsection{Developing \mo, an LLM-based Medical Conversational Agent}
\label{subsec:developing mo}

\subsubsection*{Objective}
The objective of Alan’s conversational AI agent, called \mo, is to provide users (i.e., patients) with clear, appropriate, and actionable responses to their medical and healthcare questions. Achieving this objective requires the agent to effectively acquire information from the user, analyze the information, and formulate a reliable response and recommendation grounded in sound medical knowledge and reasoning - all while maintaining positive rapport and trust. 

\subsubsection*{A Multi-Agent Aystemic Approach}
Rather than a single, standalone LLM, the agent behind \mo\ is an LLM-based AI system, consisting of several sub-agents (i.e., LLMs) that run in parallel. This multi-agent systemic approach allows \mo\ to use the best model for each specific task, integrating the strengths of different models within the system \cite{rasal2024navigatingcomplexityorchestratedproblem,guo2024largelanguagemodelbased,shen2024smallllmsweaktool}. Multi-agent systems are particularly relevant for tasks requiring deep, specialized knowledge of multiple domains as well as high accuracy and performance, as is characteristic of medicine and healthcare. 

Using a multi-agent development framework, \mo\ leverages several models initially developed by OpenAI, Anthropic, and Mistral AI. The models are served by Microsoft Azure and Google Cloud Platform (GCP) in compliance with EU privacy regulations and French health data protection requirements (HDS certification). Leveraging the existing capabilities of these models for healthcare applications requires extensive tailoring and optimization. A robust evaluation process determines which models perform best for each task and under which circumstances. 
 
\subsubsection*{Design Process and Offline Evaluation}
To design \mo's AI system architecture and select its constituent LLMs, we developed a comprehensive offline evaluation framework. The selection process for individual models was guided by core capabilities: medical knowledge, reasoning, and communication style, alongside operational requirements of speed, privacy compliance, and available capacity. We developed three critical assets for offline evaluation: (i) a clinical knowledge and reasoning benchmark, (ii) anonymized past conversations from the medical advice chat, and (iii) simulated conversations with patient agents.

\paragraph{Clinical knowledge and reasoning benchmark.}
To evaluate single models on medical knowledge and clinical reasoning, we developed a benchmark focused on French medical practice and guidelines. We extracted 800+ multiple-answer closed questions from the French national exam used to match medical school graduates to residency programs and specialties. We submitted all models to this benchmark and used their performance to inform whether and how to use them in the larger AI system.

\begin{figure}[H]
    \centering    
    \includegraphics[width=\textwidth]{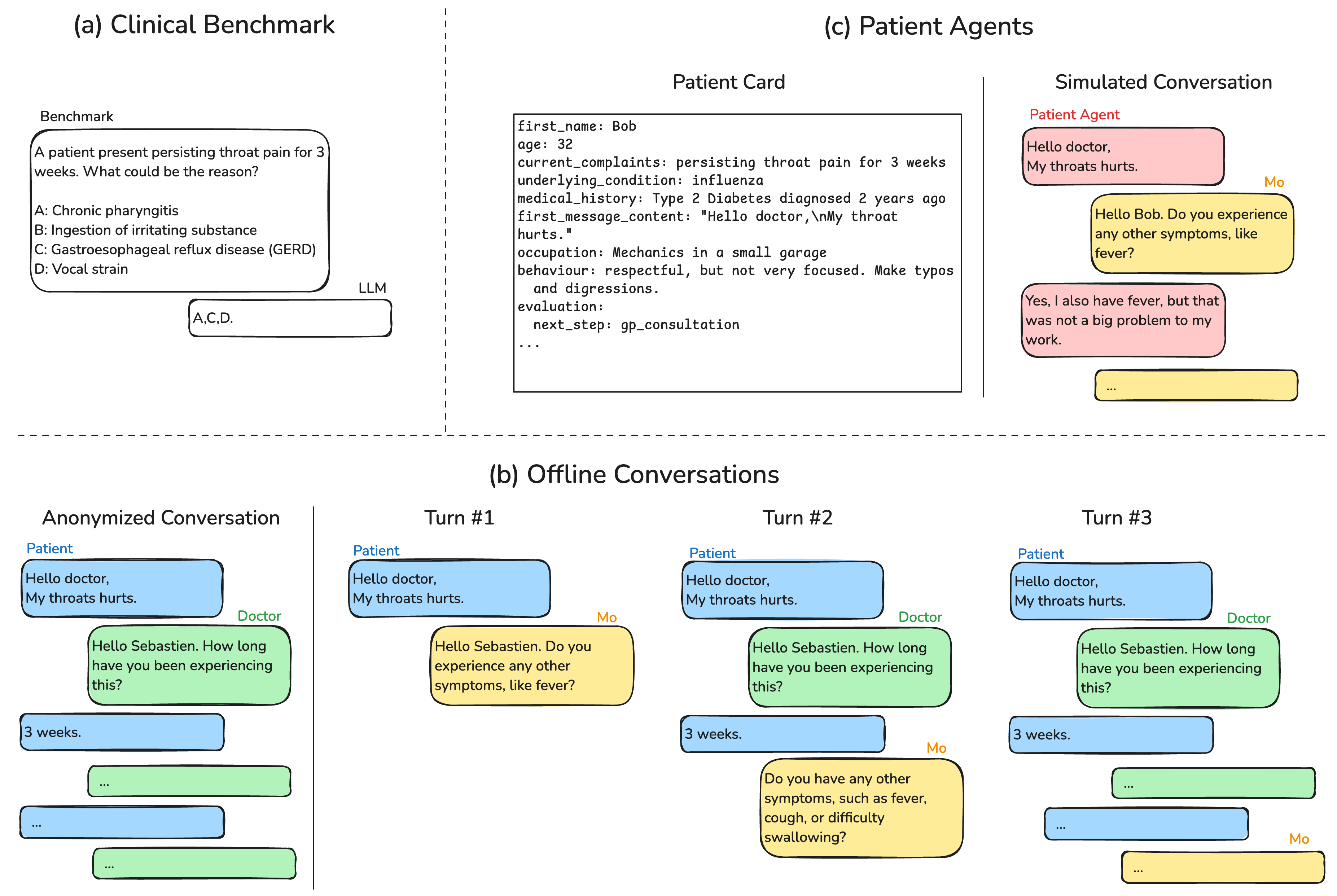}
    \caption{\textbf{Offline evaluation methods.} \textbf{(a)} Multiple-choice medical exam questions assess French medical knowledge and clinical reasoning. \textbf{(b)} Real-world medical advice conversations evaluate response quality and relevance. \textbf{(c)} Simulated conversations with patient agents evaluate end-to-end information gathering and recommendation accuracy. }
    \label{fig:offline_evaluation_methods}
\end{figure}

\paragraph{Real-world medical advice conversations.}
The agent’s goal is to provide reliable and informed replies to patients’ questions. To test this, we curated a proprietary dataset of anonymized conversations conducted on Alan’s medical advice chat service.
We truncated dialogues at points where a GP was expected to respond and submitted the unfinished conversations to the agent to test its subsequent response (see Figure \ref{fig:offline_evaluation_methods}). A physician reviewed the agent’s proposed messages to evaluate behavior, tone, and content accuracy at specific points in the conversation. While this method effectively assessed the quality of individual responses, it couldn’t capture the agent’s ability to drive full conversations independently. In particular, it didn't evaluate how well the agent could proactively gather the information needed to make sound medical assessments and recommendations.

\paragraph{Simulated conversations with patient agents.}
To address this limitation, we developed a method to evaluate complete end-to-end conversations between patients and the agent. We implemented a separate LLM-based agent designed to emulate patients in chat conversations (see Figure \ref{fig:offline_evaluation_methods}). This patient agent operates based on “patient cards”: structured inputs that define the simulated patient’s demographic characteristics, medical history, underlying medical condition and contextual information. In order to represent a range of patient communication styles and personalities, the patient card also directed how the simulated patient should behave during the exchange.
This allowed evaluation of the agent in an end-to-end setup that closely mimicked reality. Simulated conversations assessed the agent’s ability to gather relevant information, drive the dialogue, and issue reliable and appropriate recommendations. This approach also allowed us to over-represent rare or yet unseen cases, thereby evaluating the agent’s behavior in difficult scenarios and a wide range of emergency situations.

While comprehensive offline evaluation provided the foundation for safe initial deployment, evaluation using real-world data remains essential both for ensuring continued operational safety and for enabling improvements based on actual patient interactions.

\subsection{Integrating \mo\ into the medical advice chat service}
\label{subsec:integrating mo}
A product team of engineers, designers, doctors, and user researchers collaborated to integrate \mo\ into the medical advice chat service in a safe, intuitive, and transparent way. \mo\ was deployed between 9 am and 11 pm for conversations addressed to GPs in France, with patients who consented to automated treatment of their data.

\subsubsection*{Ethical Compliance}
We established comprehensive guidelines to ensure ethical compliance. We anticipated the entry into force of the EU AI Act \cite{EUAIAct2024}, augmenting its recommendations to ensure responsible implementation and a transparent interface that patients can easily understand.

\begin{figure}[H]
    \centering
    \begin{minipage}[t]{0.65\textwidth}
        \begin{subfigure}[b]{0.48\textwidth}
            \centering
            \includegraphics[width=\textwidth]{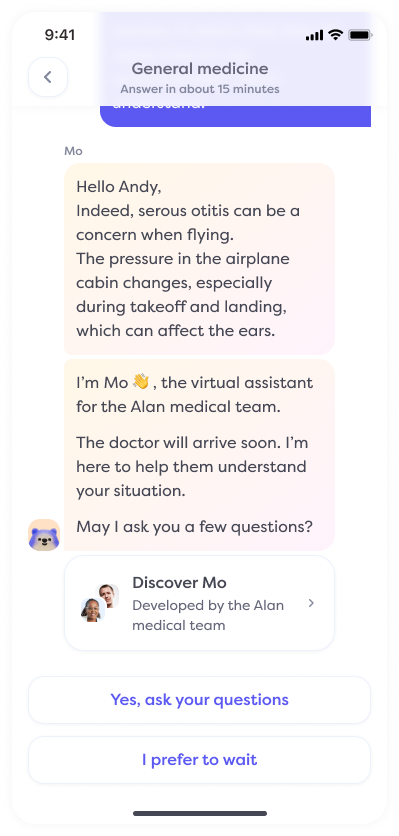}
            \caption{\textbf{Explicit start with \mo}}
            \label{fig:patient_mo_choice}
        \end{subfigure}
        \hfill
        \begin{subfigure}[b]{0.48\textwidth}
            \centering
            \includegraphics[width=\textwidth]{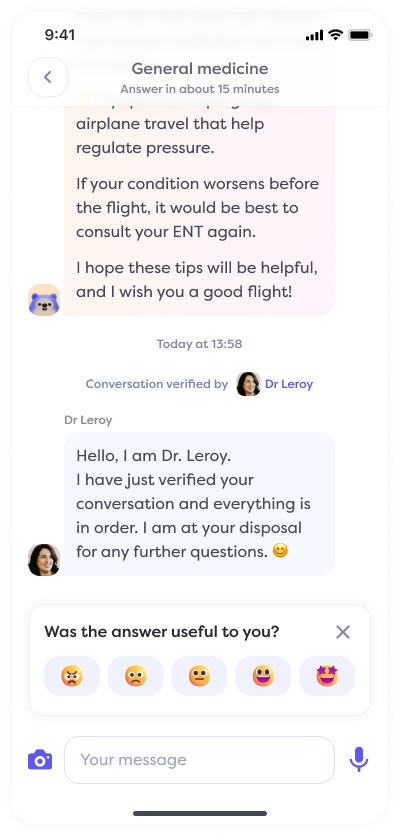}
            \caption{\textbf{Physician oversight}}
            \label{fig:dr_reviewing_mo}
        \end{subfigure}
    \end{minipage}
    \hfill
    \begin{minipage}[t]{0.32\textwidth}
        \vspace{-29.5\baselineskip} 
        \caption{
            \textbf{Transparent user interface}\\[1ex]
            \textbf{(a)} When patients initiate a conversation in the medical advice chat, \mo\ first reformulates their concern and explicitly asks for their preference: they can either start with \mo's assistance or opt to wait for a physician.\\[1ex]
            \textbf{(b)} At the end of \mo\ interactions, physicians engage directly with the patient to acknowledge their oversight of the conversation, validate \mo's medical guidance, and provide complementary advice when necessary. Here, we also show the entry point for the user ratings survey.
        }
        \label{fig:transparent_user_interface}
    \end{minipage}
\end{figure}

To ensure responsible AI deployment, we implemented the following safeguards: (1) timely human review consisting in physician oversight (2) explicit and implicit (e.g., color of text bubbles) differentiation between AI agents and human actors, (3) consent collection for health data processing using LLMs, (4) requiring positive action for interaction with \mo\ (see Figure \ref{fig:dr_reviewing_mo}), and (5) clearly limiting the scope of conversations for which \mo\ can operate. For example, in cases of psychological emergency, \mo\ was inactivated.

\subsubsection*{Physician Oversight}
\mo\ operates under the supervision and responsibility of the physicians of the medical advice chat service.

\paragraph{Physician-agent interface.} GPs have the authority and capability to stop \mo\ and intervene during any patient-agent conversation, regardless of whether \mo\ is composing a message or waiting for the patient to reply. \mo\ never resumes the conversation once stopped. The GP is required to check in with the patient after the exchange between \mo\ and the patient is complete.

\paragraph{Message review.} As a conversation between a patient and \mo\ unfolds, a GP assigned to the conversation is required to review each message from \mo\ within 15 minutes. GPs can hide \mo’s messages when necessary. Hiding a message requires the GP to take over the discussion, and displays the message in a “hidden” state to the patient while keeping it visible to the GP. In cases of urgency, GPs can immediately establish direct contact with patients using their provided contact information. 

\begin{figure}[H]
    \centering
    \begin{minipage}[c]{0.55\textwidth}
        \centering
        \includegraphics[width=0.7\textwidth]{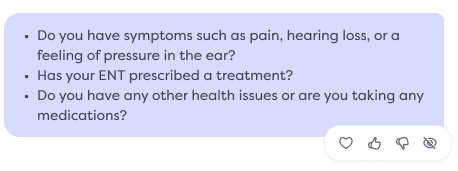}
    \end{minipage}%
    \hfill
    \begin{minipage}[t]{0.45\textwidth}
        \vspace{-3\baselineskip}
        \caption{
            \textbf{Physician review interface for \mo\ messages}.\\Physicians review each \mo\ message and select one of the four rating icons within 15 minutes. The right-most choice removes the message from the patient’s view.
        }
        \label{fig:mo_message}
    \end{minipage}
\end{figure}


\paragraph{General conversation review.} If \mo\ has been involved in a conversation, the assigned GP must perform a general review. This review consists of examining the complete \mo-patient dialogue to evaluate the medical advice provided and identify any potential gaps or concerns. The GP then documents their assessment and engages directly with the patient for a mandatory check-in to confirm their oversight, validate \mo's medical recommendations, provide complementary guidance when needed, and address any remaining questions (see Figure \ref{fig:dr_reviewing_mo}).

\subsubsection*{Staged Roll-out and Quality Assurance}
\label{Staged Rollout}

\mo's deployment progressed through three sequential stages over a four-month period ending in October 2024. The first stage limited access to Alan employees only, allowing for initial validation. The service was then extended to a small proportion of Alan members under the supervision of GPs selected and trained to support \mo’s development. Finally, access was expanded to 50\% of members with oversight from all GPs of the medical advice chat service after they received specific training. Each stage lasted as long as necessary to reach defined safety and stability milestones.

Throughout the integration, a team of physicians and engineers continuously monitored safety and stability metrics established during development, enabling data-driven improvements while maintaining rigorous quality standards.

\section{Methods}
\label{sec:methods}

\subsection{Study Design}
\label{subsec:study design}
We conducted a randomized controlled experiment to evaluate the effect of \mo, our LLM-based conversational agent, on patient experience. Of all conversations where \mo\ was activated, only those considered in scope were eligible to have \mo\ engage with the patient. From this pool of eligible conversations, \mo\ was proposed to a random 50\% sample of patients to  comprise the treatment group. The remaining eligible conversations, where \mo\ was not proposed, served as the control group. We evaluated patient experience across three domains: (i) overall satisfaction, (ii) quality metrics (clarity, trust, and empathy), and (iii) engagement metrics (response patterns).

In addition to assessing patient experience, we evaluated \mo’s safety and medical accuracy from the physician message and general conversation reviews.

Data was prospectively collected from September 30 to October 20, 2024.

\subsection{Outcome measures}
\label{subsec:outcome measures}

We developed questionnaires to evaluate both the patient experience of conversations with \mo\ and the physician assessments of safety and accuracy of \mo’s responses. To do so, we surveyed existing standards for evaluation of patient-doctor interactions (PACES exam \cite{PACES2023}, GMC Patient Questionnaire \cite{GMCPatientFeedback}, Best Practice for Patient Centered Care \cite{King2013})  and extracted core information on our specific domains of interest. We differentiated between patient-related outcomes to be reported by the patient and medical assessment to be conducted by a physician, while considering constraints in length and user experience to maximize completion rate. 

\subsubsection*{Patient Ratings}
Following each conversation, patients were asked to rate their experience across four dimensions: overall satisfaction, clarity, trust, and empathy (see Table \ref{tab:patient_feedback_table}, Supplementary Figure \ref{fig:user_screens}). Information on patient satisfaction was captured using a 5-point Likert scale and free text. Clarity, trust, and empathy were assessed using a 4-point Likert scale.

\begin{table}[H]
    \caption{\textbf{Patient experience questionnaire}}
    \vspace{4pt}  
    \label{tab:patient_feedback_table}
    \fontsize{8pt}{9pt}\selectfont
    \renewcommand{\arraystretch}{2}
    \begin{tabular*}{\textwidth}{@{\extracolsep{\fill}}p{0.15\textwidth}p{0.38\textwidth}p{0.37\textwidth}}
        \toprule
        \fontsize{10pt}{10pt}\selectfont{Category}              & \fontsize{10pt}{10pt}\selectfont{Question}                                               & \fontsize{10pt}{10pt}\selectfont{Rating Scale} \\
        \midrule
        Overall satisfaction  & How useful was the conversation?                       & \raisebox{-.3\height}{\includegraphics[height=2em]{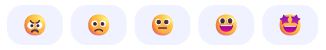}} \\
        Clarity               & How clear was the information you've received?         & Not at all; Not very; Substantially; Perfectly \\
        Trust                 & How much do you trust the information you've received? & Not at all; Not very; Substantially; Perfectly \\
        Empathy               & How heard and understood have you felt?                & Not at all; Not very; Substantially; Perfectly \\
        \bottomrule
    \end{tabular*}
\end{table}

\begin{table}[H]
    \fontsize{8pt}{9pt}\selectfont
    \renewcommand{\arraystretch}{2}
    \caption{\textbf{General practitioner assessment}}
    \vspace{4pt}  
    \begin{tabular*}{\textwidth}{@{\extracolsep{\fill}}p{0.15\textwidth}p{0.38\textwidth}p{0.37\textwidth}}
        \toprule
        \fontsize{10pt}{10pt}\selectfont{Category} & \fontsize{10pt}{10pt}\selectfont{Question} & \fontsize{10pt}{10pt}\selectfont{Assessment} \\
        \midrule\addlinespace[2.5pt]
        Advice & Are \mo's recommendations clear and appropriate? 
            & \textbf{Dangerous:} Wrong advice, potentially dangerous \newline
              \textbf{Insufficient:} Not very clear, not very actionable, or not well-suited to the patient's needs \newline
              \textbf{Good:} Sufficiently clear, actionable and suitable \newline
              \textbf{Excellent:} Impressive by some aspects \\
       
        Questions & Are \mo's questions relevant and well-phrased? 
            & \textbf{Dangerous miss:} Essential questions are missing or poorly phrased \newline
              \textbf{Insufficient:} Some missing or poorly phrased questions \newline
              \textbf{Good:} Sufficient questions posed \newline
              \textbf{Excellent:} Perfect! No unnecessary questions. \\
    
        Accuracy & Do \mo's messages contain inaccuracies or confabulations? 
            & \textbf{Dangerous errors:} Potentially dangerous inaccuracies or confabulations \newline
              \textbf{Yes:} Inaccuracies or confabulations without danger \newline
              \textbf{No:} No inaccuracy or confabulation. \\
    
        Overall Assessment & Overall, the conversation between \mo\ and the patient seemed to you...
            & \textbf{Dangerous} \newline
              \textbf{Laborious} \newline
              \textbf{Satisfactory} \newline
              \textbf{Amazing} \\
        \bottomrule
    \end{tabular*}
    \label{tab:gp_feedback_table}
\end{table}

\subsubsection*{GP General Review}
After each complete \mo-patient conversation, the assigned GP evaluated its quality. They assessed \mo's questioning, recommendations, and accuracy, and also provided an overall assessment of the conversation. All used a 4-point Likert scale apart from accuracy, which was rated on a 3-level scale (see Table \ref{tab:gp_feedback_table}, Supplementary Figure \ref{fig:gp_screens}).

\subsubsection*{Statistical Analysis}
We compared distributions of patient and GP ratings using the Wilcoxon test. Demographic comparisons were conducted using Student’s t-test for age and chi-squared test for gender.

We excluded from the study all conversations with attachments (document, picture) and conversations with Alan employees.

Data from conversations requesting unavailable services (prescriptions, sick leave certificates, or medical certificates) were excluded from the patient experience analysis.

All statistical analyses were conducted using R version 4.3.1. 

\subsubsection*{Data Privacy and Consent for Research Use}
All members included in this study were informed of the use of aggregated and/or anonymized data for research and statistical purposes in Alan’s Privacy Policy. This privacy policy specifies that data collected by Alan may be utilized for scientific research in a manner compatible with the original purpose of collection, ensuring that all data analyzed remains non-identifiable and protects individual privacy. Additionally, members who used this specific service provided explicit consent through a dedicated consent screen for the automated processing of their health data using LLM technology.

\section{Results}
\label{sec:results}
\subsection{Sample Profile}
\label{subsec:sample profile}

\begin{figure}[H]
    \centering
    \begin{minipage}[t]{0.45\textwidth}
        \centering
        \includegraphics[width=0.67\textwidth]{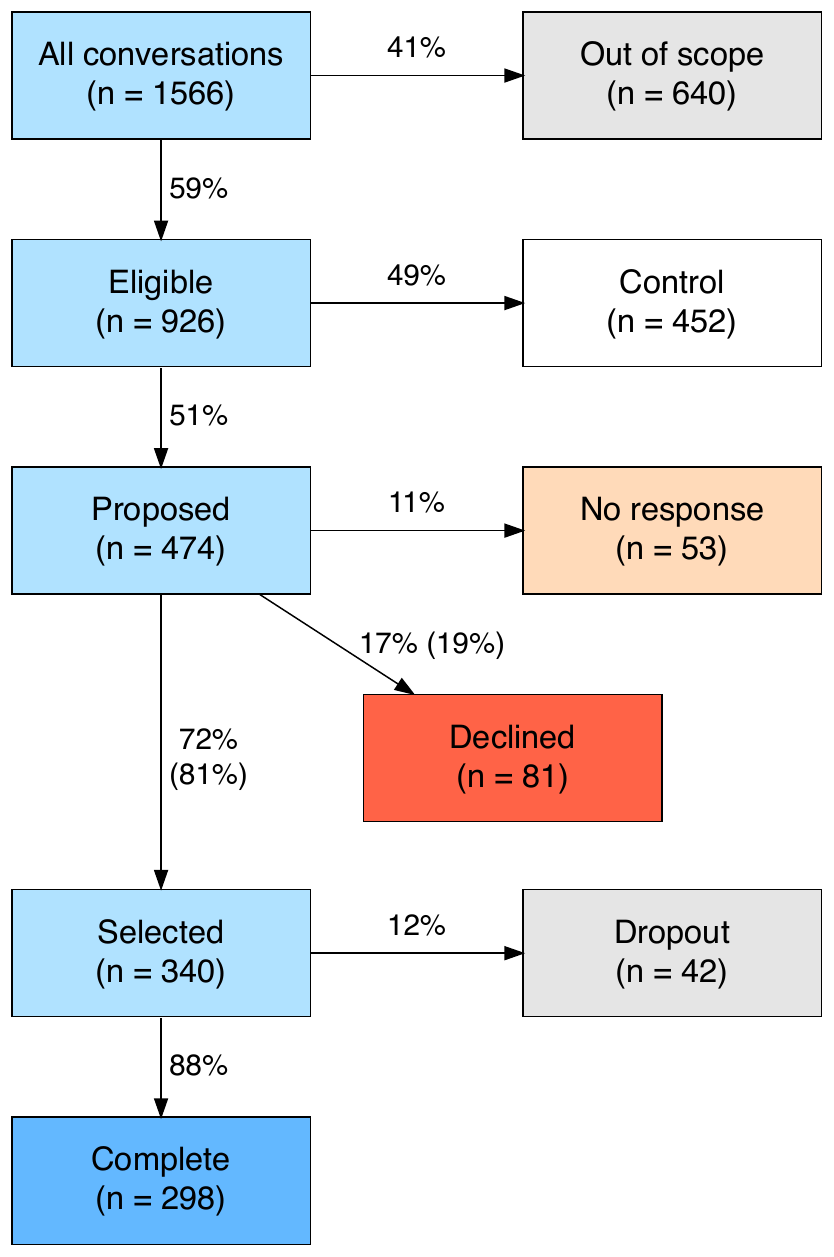}
    \end{minipage}
    \begin{minipage}[t]{0.45\textwidth}
        \vspace{-19.5\baselineskip} 
        \caption{\textbf{Flow diagram of \mo\ deployment in medical advice conversations.} Of 1,566 conversations where \mo\ was active, 640 (41\%) were out of scope. Among eligible conversations (n = 926), \mo\ was proposed to 474 patients, with 452 as controls. After excluding no-responses (n = 53) and declines (n = 81), 340 patients opted to interact with \mo, of whom 298 (88\%) completed their conversations. Percentages in parentheses represent rates adjusted for no-responses.}
        \label{fig:deployment_flow_diagram}
    \end{minipage}
\end{figure}

Over the study period, 1,566 conversations were initiated in Alan's medical advice chat service during \mo's active hours (Figure \ref{fig:deployment_flow_diagram}). \mo\ deemed 640 conversations (41\%) out of scope, due to questions that contained insurance or administrative matters or signs of mental health distress that, by established protocols, required human intervention.

Of the 926 eligible conversations, \mo\ was proposed to 474 patients (51\%), while 452 conversations served as the control group. Among those offered \mo, 53 patients (11\%) did not respond within the required 15-minute window before GP takeover, likely because they expected an asynchronous response and were not actively monitoring their chat. Of the remaining patients who responded, 81 (19\%) declined interaction, resulting in 340 patients opting to interact with \mo, an acceptance rate of 81\% among respondents.

Among those who began interacting with \mo, 298 patients (88\%) completed their conversations, while 42 patients (12\%) dropped out before completion as assessed by the monitoring physician.

\begin{table}[H]
\fontsize{8pt}{9pt}\selectfont
\caption{\textbf{Demographic characteristics by conversation status and group}\\Age and gender distribution across conversation categories. Age is presented as mean and range [25th - 75th percentiles], with minimum and maximum values. F prop. represents the proportion of conversations with female users. Groups are mutually exclusive and follow the flow diagram (Figure \ref{fig:deployment_flow_diagram}).}
\begin{tabular*}{\linewidth}{@{\extracolsep{\fill}}lrrcrrr}
                                        &               & \multicolumn{2}{c}{Age}      & \multicolumn{3}{c}{Gender} \\ 
\cmidrule(lr){3-4} \cmidrule(lr){5-7}
                                        & Conversations & Mean   & min [q25 - q75] max & Female & Male  & F prop. \\ 
\midrule\addlinespace[2.5pt]
All Conversations                       & 1,566         & 34.5   & 17 [28 - 39] 72     & 983    & 575   & 63\% \\ 
{\bfseries Eligible}                    & \textbf{926}         & \textbf{32.1} & \textbf{18 [27 - 36] 67}   & \textbf{619}  & \textbf{302} & \textbf{67\%} \\ 
\midrule\addlinespace[2.5pt]
{\hspace{11.25pt}Control}               & 452           & 31.9   & 18 [27 - 35] 67     & 304    & 146   & 68\% \\ 
{\bfseries \hspace{11.25pt}\mo\ Proposed} & \textbf{474}         & \textbf{32.3} & \textbf{18 [27 - 36] 64}   & \textbf{315}  & \textbf{156} & \textbf{67\%} \\ 
\midrule\addlinespace[2.5pt]
{\hspace{22.5pt}\mo\ No Answer}           & 53            & 33.4   & 18 [29 - 36] 64     & 34     & 19    & 64\% \\ 
{\hspace{22.5pt}\mo\ Declined}            & 81            & 31.0   & 18 [26 - 34] 55     & 48     & 32    & 60\% \\ 
{\bfseries \hspace{22.5pt}\mo\ Selected}  & \textbf{340}         & \textbf{32.5} & \textbf{18 [27 - 36] 63}   & \textbf{233}  & \textbf{105} & \textbf{69\%} \\ 
\midrule\addlinespace[2.5pt]
{\hspace{33.75pt}Dropout}               & 42            & 31.7   & 20 [26 - 36] 53     & 29     & 11    & 72\% \\ 
{\bfseries \hspace{33.75pt}Complete}    & \textbf{298}         & \textbf{32.6} & \textbf{18 [27 - 36] 63}   & \textbf{204}  & \textbf{94}  & \textbf{68\%} \\ 
\bottomrule
\label{tab:table_demographics}
\end{tabular*}
\end{table}

The demographic characteristics across conversation categories are presented in Table \ref{tab:table_demographics}. The mean age of users across all conversations was 34.5 years, with a higher proportion of female users (63\%). Among eligible conversations, the control and \mo\  Proposed groups showed comparable demographic profiles (mean age difference: 0.4 years [95\% CI: -0.5 to 1.4]; difference in female proportion: -0.7\% [95\% CI: -7.0\% to 5.6\%]). The demographic characteristics in completed conversations (mean age: 32.6 years, 68\% female) remained consistent with the initial eligible population (mean age difference: 0.5 years [95\% CI: -0.6 to 1.5]; difference in female proportion: 1.3\% [95\% CI: -5.0\% to 7.6\%]).

\subsection{Patient Experience}
\label{subsec:patient experience}

Patient ratings were available for 20\% of eligible conversations. Ratings were more prevalent in the control group (24\% vs 17\%), and demographic characteristics were comparable between the two groups (mean age difference: 1.6 years [95\% CI: -0.8 to 3.9]; difference in female proportion: -3\% [95\% CI: 11\% to 17\%]).

\begin{figure}[H]
    \centering
    \begin{subfigure}{\textwidth}
        \centering
        \includegraphics[width=0.8\textwidth]{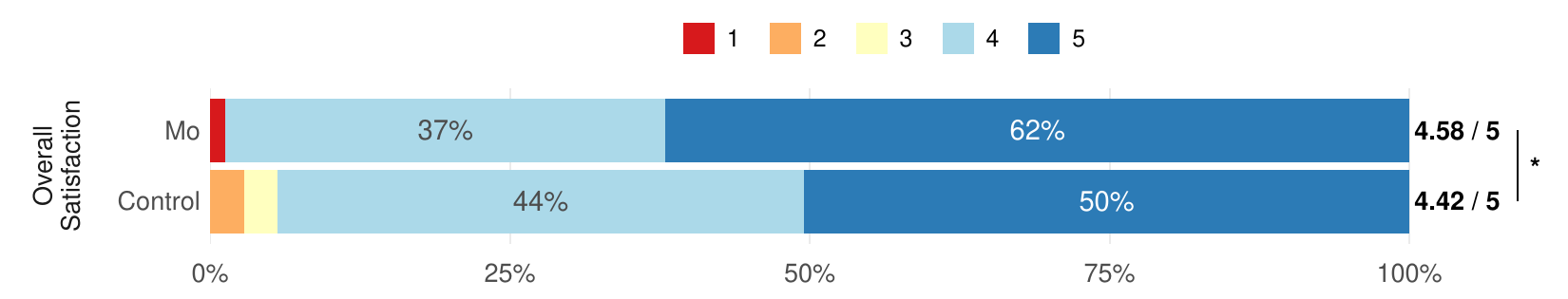}
    \end{subfigure}
    
    \vspace{1em}
    
    \begin{subfigure}{\textwidth}
        \centering
        \includegraphics[width=0.8\textwidth]{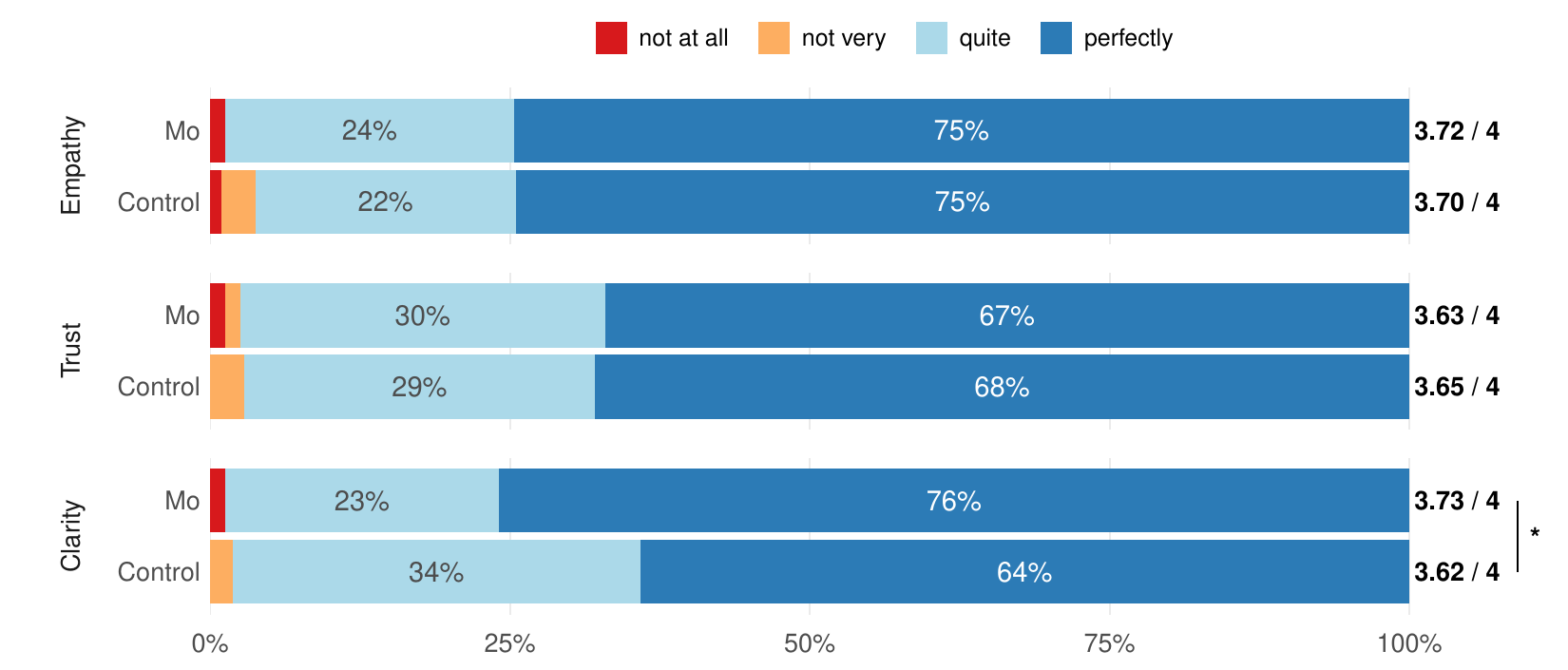}
    \end{subfigure}
    
    \caption{\textbf{Patient ratings: comparison between \mo\ and control groups.} Distribution of patient ratings for \mo\ and control groups across different dimensions. \textbf{Top:} Overall satisfaction rated on a 5-point scale (1:\raisebox{-.2em}{\includegraphics[height=1em]{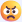}}, 5:\raisebox{-.2em}{\includegraphics[height=1em]{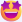}}). \textbf{Bottom:} Specific dimensions (Empathy, Trust, Clarity) rated on a 4-point scale ('not at all' to 'perfectly'). Numbers on the right show mean scores. Asterisks (*) indicate statistically significant differences between groups (p < 0.05). Percentages show proportions of responses in each category.}
    \label{fig:patient_ratings}
\end{figure}

\mo\ received higher general satisfaction scores compared to the control group (mean: 4.58 vs 4.42 out of 5, p < 0.05) (Figure \ref{fig:patient_ratings}). Both treatment and control groups showed similar ratings for trust (mean: 3.63 vs 3.65 out of 4) and empathy (mean: 3.72 vs 3.70 out of 4). However, \mo\ achieved significantly higher clarity ratings (mean: 3.73 vs 3.62 out of 4, p < 0.05). 

Notably, extremely low ratings (score of 1) were rare. \mo\ received only one such rating across all dimensions, and the control group received one rating of 1 for empathy only. A detailed analysis of all ratings below 3 (n = 8) revealed no systematic patterns of dissatisfaction (Supplementary Table \ref{tab:conversation_details}).

\subsection{Patient Engagement}
\label{subsec:patient engagement}
\begin{figure}[H]
    \centering
    \begin{minipage}[t]{0.65\textwidth}
        \centering \includegraphics[width=\textwidth]{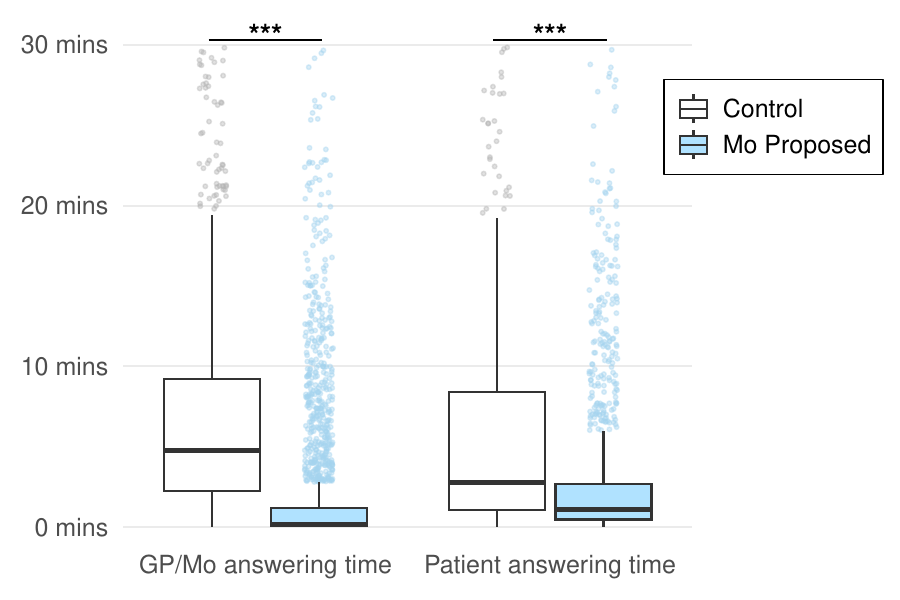}
    \end{minipage}
    \hfill
    \begin{minipage}[t]{0.34\textwidth}
        \vspace{-18\baselineskip} 
        \caption{
            \textbf{Response time distributions in medical chat conversations}\\[1ex]
            \textbf{Left:} Time taken by providers to respond (\mo\ or GP) after the patient.\\[1ex]
            \textbf{Right:} Time taken by patients to respond.\\[1ex]
            Box plots show median, interquartile range, and whiskers (1.5 IQR); individual points represent outliers beyond whiskers. The visualization is cropped on the Y axis. In the \mo\ Proposed group, patients interact with both \mo\ and the GP. Asterisks (***) indicate statistically significant differences (p < 0.001).
        }
        \label{fig:response_time_distribution}
    \end{minipage}
\end{figure}

We analyzed conversation dynamics by measuring response times for each turn of dialogue between participants (Figure \ref{fig:response_time_distribution}). In the control group, these turns were exclusively between patients and GPs, while in the \mo\ group, turns included both \mo-patient and GP-patient interactions.

As expected, since \mo\ responds almost instantaneously, response times from providers differed significantly (median: 0.2 vs 4.8 minutes, p < 0.001). Interestingly, this difference in provider response times was accompanied by a change in patient behavior: in conversations with \mo, patients also responded more quickly compared to control conversations (median: 1.1 vs 2.8 minutes, p < 0.001).

\subsection{Safety and Medical Accuracy}
\label{subsec:safety and medical accuracy}
\begin{figure}[H]
    \centering
    \begin{subfigure}{\textwidth}
        \centering
        \includegraphics[width=0.7\textwidth]{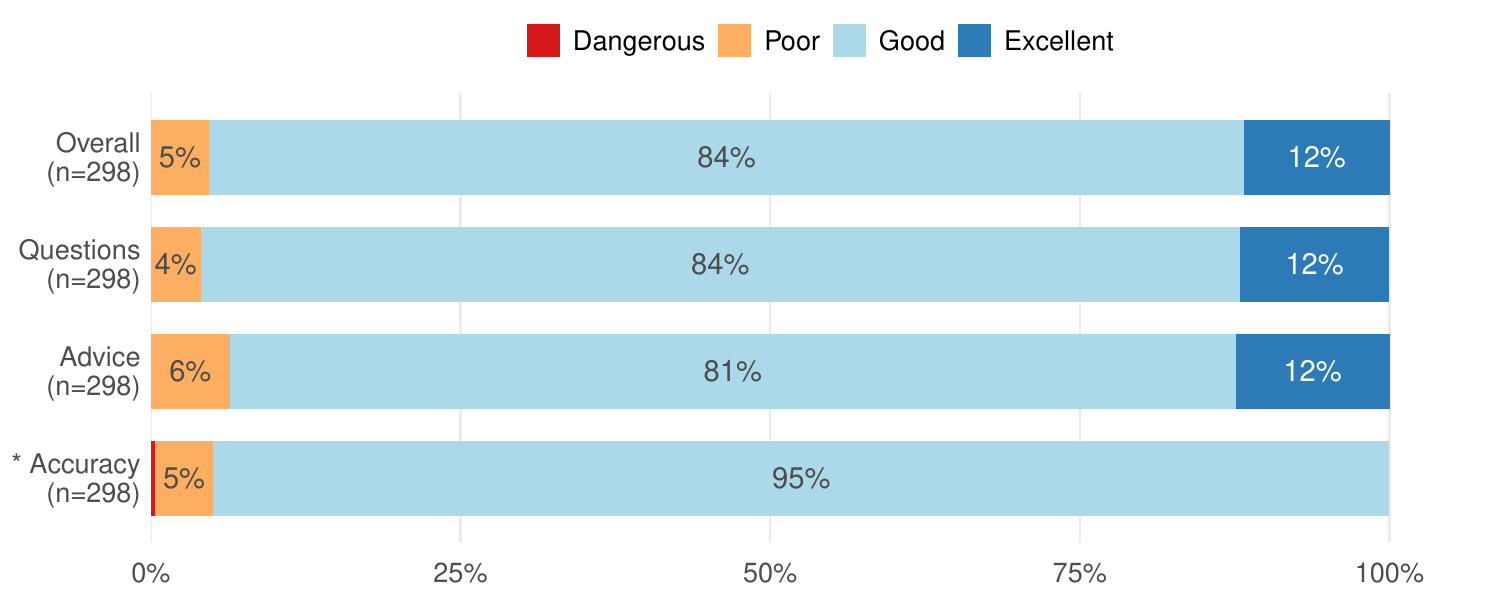}
    \end{subfigure}
    \begin{subfigure}{\textwidth}
        \centering
        \includegraphics[width=0.7\textwidth]{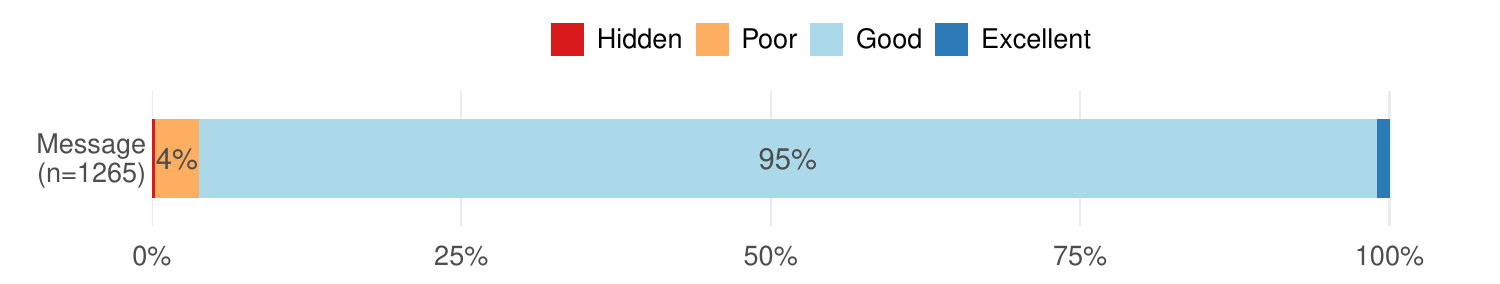}
    \end{subfigure}
    \caption{\textbf{GP evaluation of \mo's medical quality at message and conversation levels.} \textbf{Top:} Conversation-level assessment (n=298) across different dimensions. Each conversation was evaluated for overall performance, quality of questions asked, advice given, and accuracy. Ratings range from ``dangerous'' (red) to ``excellent'' (dark blue), except for Accuracy (*) which was rated specifically for presence of inaccuracies (none/some/dangerous). \textbf{Bottom:} Message-level review (n=1,265) of individual responses from \mo, rated from ``hidden'' (red) to ``excellent'' (dark blue).}
    \label{fig:gp_ratings}
\end{figure}

GPs supervising the medical advice chat service evaluated \mo's performance at both message and conversation levels (Figure \ref{fig:gp_ratings}). At the message level, supervising GPs reviewed each of \mo's responses within 15 minutes of sending. Among 1,265 messages sent by \mo, 95\% were rated positively, while 45 messages (3.6\%) were rated as “poor” and 3 messages were hidden from patients. No harm resulted from the messages that were subsequently hidden from patient view.

Following the completion of each conversation, GPs provided an overall assessment. For completed conversations (n=298), 95\% received positive ratings (“good” or “excellent”) for overall performance, with similar distributions for question quality (96\%) and advice appropriateness (94\%). No conversation was deemed potentially dangerous overall.

In the assessment of medical accuracy, 95\% of conversations contained no inaccuracies, with one conversation flagged for the presence of potentially dangerous inaccuracies.

\section{Discussion}
\label{sec:discussion}
This study presents the first large-scale evaluation of a physician-supervised LLM-based conversational agent in a real-world medical setting. By integrating \mo\ into an existing medical advice chat service, we demonstrated that AI-assisted conversations achieved comparable or superior patient experience while maintaining robust safety standards under physician oversight. Notably, patients reported higher information clarity and overall satisfaction when interacting with \mo\ compared to standard care, while showing equivalent levels of trust and perceived empathy. General practitioners with extensive experience in medical chat services assessed 95\% of \mo's conversations as good or excellent. Together, these findings from both patients and physicians suggest strong potential for AI augmentation in healthcare communication.

\subsection{Bridging AI Research and Clinical Practice}
\label{subsec:bridging ai research and clinical practice}
The transition from AI research to clinical implementation represents a critical frontier in healthcare innovation. This section examines the current landscape and contextualizes our contributions within existing literature.

\subsubsection*{Evaluation of Large Language Models tailored to the medical field}
Substantial effort has focused on developing and evaluating LLMs specifically trained for the health domain (e.g., MedPalm \cite{singhal2023expertlevelmedicalquestionanswering}, DrBert \cite{labrak2023drbertrobustpretrainedmodel}). While these studies demonstrated promising capabilities on medical knowledge benchmarks, their evaluations primarily employed objective closed-question assessments that do not fully capture the complexities of patient interactions. In a related direction, Ayers et al. (2023) retrospectively demonstrated superior quality and empathy of LLM responses compared to those coming from physicians on a public forum, though this baseline may not reflect professional medical care \cite{Ayers2023}.

\subsubsection*{AI-driven Clinical Decision-Making}
The development of large-scale symptom assessment systems for disease diagnosis and patient triage marks a significant advancement in AI-driven healthcare. The large study (n=102,059) of Zeltzer et al. (2023) demonstrated the potential for AI to enhance primary care triage \cite{Zeltzer2023}. However, these systems typically operate within narrowly defined parameters of  structured symptom assessment, leaving unexplored the broader range of medical queries that arise in primary care settings.

The research conducted by Hager et al. (2024), analyzing 2,400 cases of abdominal pathology, revealed that LLMs had notably lower diagnostic accuracy compared to human physicians \cite{Hager2024}. Although newer proprietary LLM versions and multi-agent systems might improve these results, their findings advocate for a supervised integration of LLMs in clinical practice (healthcare professional oversight, continuous validation, ongoing research) as complementary tools rather than fully autonomous systems.

\subsubsection*{Simulated Clinical Interactions}
The AMIE system represents a major step forward in patient-facing medical AI, showing superior diagnostic accuracy and performance in clinical dialogue \cite{tu2024conversationaldiagnosticai}. Their robust evaluation framework, including a double-blind comparison with physicians, provides valuable insights. However, the study’s limitations should be noted: it was conducted in a simulated environment with patient actors, and the participating physicians were new to chat-based consultations, potentially failing to reflect the expertise of clinicians experienced in digital healthcare delivery.

\subsubsection*{Limited-Scale Real-World Applications}
Several studies have explored real-world deployments of conversational AI agents in specific healthcare contexts, including postoperative recovery (\cite{Dwyer2023}, n=26), older adult patient-provider communication (\cite{Yang2024}, n=19), and loneliness mitigation (\cite{Jo2023}, n=34). While these studies consistently report improved patient satisfaction and reduced provider workload, their limited sample sizes constrain broader generalization.

\subsection{Understanding Patient Experience: Satisfaction, Trust, and Engagement}
\label{subsec:satisfaction trust and engagement}

\subsubsection*{Implications for Healthcare Delivery}
Building on these promising but limited pilots, our study presents the first large-scale deployment of an AI medical assistant in a real-world healthcare setting, with close to 300 completed patient conversations. Our findings on patient satisfaction merit careful interpretation within the broader context of healthcare delivery. Patient satisfaction is a crucial prerequisite for broader acceptance and adoption of AI in healthcare. The comparable or superior satisfaction ratings achieved in conversations with \mo\ indicates the feasibility of AI deployment in clinical settings. This acceptance could enable significant reconfiguration of healthcare delivery systems, potentially allowing for more efficient allocation of human medical expertise while maintaining or improving access to care. Specifically, AI agents could evolve into daily health companions, fundamentally shifting healthcare from episodic interventions to continuous support, where patients are empowered to better understand and manage their health journey, while being efficiently connected to physician expertise when needed.

\subsubsection*{Dimensions of Patient Satisfaction}
The granular analysis of satisfaction metrics reveals important nuances in patient experience. The significantly higher clarity ratings suggest that AI-assisted communications may excel at providing clear, structured information, aligning with previous findings that standardized communication approaches can enhance patient understanding \cite{Trevena2005}. 

The equivalent ratings for trust and empathy warrant particular attention. Unlike studies where raters were unaware of AI involvement (e.g., \cite{tu2024conversationaldiagnosticai,Ayers2023}), our transparent setup explicitly identified \mo\  as an AI agent. Previous research on AI interactions suggests that perceived humanness increases feelings of trust and empathy \cite{LU2022107092,Hu2021}. Therefore, the comparable ratings are especially significant given that knowledge of \mo's AI status could have influenced patient expectations. Two factors likely contributed to maintaining trust despite transparent AI use: \mo's consistent responsiveness and structured communication style, and our protocol ensuring that a physician  personally engages with the patient at the end of each conversation.

\subsubsection*{Patient Engagement and Communication Dynamics}
Analysis of conversation dynamics revealed intriguing patterns in patient engagement. \mo's nearly instantaneous responses were associated with faster patient response times, suggesting more fluid and engaged conversations. Beyond mere efficiency, these accelerated exchanges could fundamentally improve healthcare delivery. Fluid dialogue leads to more comprehensive information gathering, while rapid response times could lower the barrier to seeking medical advice, encouraging patients to address health concerns earlier. The combination of AI responsiveness and physician oversight creates a new model where patients benefit from both immediate attention and expert medical judgment. This finding aligns with previous research showing that reduced response latency can enhance user engagement and satisfaction in healthcare communications \cite{Yang2024,Wu2023}.

The high opt-in rate (81\% among respondents) indicates strong patient acceptance of AI-assisted healthcare services, setting a higher benchmark for user acceptance than previously suggested in the literature \cite{Horowitz2023,Esmaeilzadeh2021}. Through user interviews, we identified three factors potentially contributing to this success: (i) members' trust in Alan, built over time (ii) an iteratively refined user experience, and (iii) an emphasis on transparency. 

These findings suggest that successful integration of AI in healthcare services depends not only on technical capabilities but also on careful attention to user experience, institutional trust, and transparent implementation practices. The results demonstrate that when properly implemented, AI-assisted healthcare services can achieve high levels of patient acceptance while maintaining high quality standards in medical communication.

\subsection{Ethical, Privacy, and Safety concerns of AI-based Communication Systems for Health}
\label{subsec:concerns of ai communication systems for health}
From a safety perspective, the results of our study are encouraging yet warrant careful consideration. While 95\% of \mo's messages received positive physician reviews and only three messages (out of 1,265) required intervention, the few cases where mitigation was required by the supervising GP confirms the need for physician oversight in this setup and continued research. In particular, extended data collection will allow observation of a broader range of rare cases that may elicit inappropriate responses from the agent.

Earlier studies emphasized several  prerequisites for deploying patient-facing AI systems in healthcare: stringent quality control measures, sufficient guardrails, adequate oversight by qualified physicians, ethical design and development,  as well as strict adherence to privacy regulations and informed consent procedures \cite{Wu2023,Busch2024.03.04.24303733,Haltaufderheide_2024,Mesk2023}. The integration of \mo\ in Alan’s medical advice chat demonstrates a practical realization of these requirements in a real-world healthcare setting.

The following steps were critical in ensuring its reliability. First, we established comprehensive offline evaluation procedures, comprising of: (i) the constitution of an internal closed-questions benchmark, tailored to the needs relevant to the deployment of the agent, and unlikely to be used in the prior training of the LLMs we use, (ii) the use of anonymized past conversation data representative of the specific task, and (iii) the development of an automated conversation evaluation framework involving  patient agents. Second, we carefully integrated the agent in the final product, insisting on (i) the thoughtful design of the interaction between the physician and the agent, prioritizing physician oversight and leveraging user experience to elicit the right actions (e.g., timely message review), and (ii) a staged rollout to enable learning and iterations before full-scale implementation.

This study was made possible by two critical aspects of our development process. First, we build upon a pre-existing medical service. Second, the agent and its integration into the patient-facing product were developed by a multidisciplinary team that included a dedicated GP, aligning with recommendations made by others \cite{zhou2024surveylargelanguagemodels}. 

\subsection{Study Limitations}
\label{subsec:study limitations}
This real-world evaluation, while providing valuable insights, has several important limitations. First, the three-week duration of our study may not capture the full range of medical presentations. Seasonal variations in health issues could be underrepresented, and longer-term patterns in patient-AI interactions remain to be explored. More importantly, this sample size, though substantial for an initial deployment, may not be sufficient to detect rare but significant safety issues that could emerge in broader medical practice.

The evaluation of patient experience was constrained by our survey response rate of 20\%. While this rate is typical for embedded product surveys, it introduces potential selection bias in our satisfaction metrics. Despite finding no significant demographic differences between respondents and non-respondents, there may be unmeasured factors influencing survey participation that correlate with patient satisfaction.

Our study scope was also limited in several practical ways. We restricted \mo's deployment to general practitioner conversations, excluding consultations with other specialists, which might present different challenges. The exclusion of conversations requiring document review or image analysis, while necessary for our initial deployment, leaves important use cases unexplored. Additionally, as the study was conducted within a single healthcare system with an established digital presence, our findings about patient acceptance may not generalize to other healthcare contexts, particularly those without pre-existing patient trust in digital services.

\subsection{Future Research Priorities}
Our study demonstrates the potential of AI-assisted medical communication, while highlighting key areas for future research.
\label{subsec:future research priorities}

\subsubsection*{Clinical Impact Studies}
Building on our initial safety and satisfaction findings, longer-term studies should examine how AI assistance affects healthcare delivery and outcomes. Critical questions include the impact on patient health-seeking behavior, the quality of preventive care, and physician workload and burnout. Particularly important is understanding how AI assistance influences the patient journey through the healthcare system, including timely specialist referrals and follow-up care.

\subsubsection*{Healthcare System Integration}
Deeper integration into healthcare workflows presents both opportunities and challenges. Research should focus on optimizing the collaboration between AI systems and healthcare professionals, establishing efficient oversight models, and developing protocols for seamless care transitions. This includes studying how AI can enhance rather than disrupt existing care pathways, and identifying best practices for maintaining quality while improving healthcare access and efficiency.

\subsubsection*{Technical Evolution}
Several technical advances could expand the system's utility in clinical practice. Integration with electronic health records would provide richer context for patient interactions, while capabilities for handling medical documents and images would enable more comprehensive care support. Continued research into improving the handling of complex medical presentations and rare conditions remains essential for reliable deployment at scale.

\newpage
\section{Conclusion}
\label{sec:conclusion}
Our findings demonstrate the feasibility and far-reaching potential of AI-assisted medical communication, while highlighting the importance of careful implementation and oversight. The success of this implementation relied heavily on the integration of medical expertise throughout development, robust privacy protections, and continuous safety monitoring. While results are promising, longer-term studies with larger sample sizes are needed to fully understand the impact of AI-assisted medical communication on healthcare delivery, access and quality of care, and patient outcomes.

\subsubsection*{Acknowledgements}
We thank the health professionals interacting with \mo\ everyday for their expertise, their flexibility and their goodwill: Ammar Alsheikhly (MD), Btissame Betari (MD), Laurène Bideau (MD), Aleksandra Culafic (MD, alpha tester), Cécile Coutant (MD), Kamyar Dadsetan (MD), Aicha Diakite (MD), Axelle Durocher (MD), Yann Kieffer (MD, medical community lead), Émilie Le Lan (MD), Adrien Leclerc (MD), Mehdi Oulmouddane (MD, alpha tester), Valeria Zuddas (MD).

We thank Joy Shi (Harvard T.H. Chan School of Public Health) for support in statistical analysis.

From Alan, we thank Hortense Villeronce (User Research) and  Francois Zannotti (Legal), for their direct contributions to the project, as well as Juan Pablo Briceno (Associate) for his support in putting this paper together.

Finally, we thank the team of Alan at large for their outstanding work over the last 8 years, without which this project would not have been possible.

\subsubsection*{Competing Interests:}
This study was funded by Alan Tech. Pierre-Auguste Beaucoté, Anaël Beaugnon, Marion Doumeingts, Antoine Lizée and James Whitbeck are employees of Alan Tech and receive stock options as part of their standard compensation package. The authors declare no other conflicts of interest.

\newpage
\bibliographystyle{unsrturl}
\bibliography{references}

\newpage
\section*{Supplementary Material}

\newcounter{supplementarytable}
\stepcounter{supplementarytable}
\renewcommand{\thetable}{S\thesupplementarytable}

\begin{table}[H]
\caption{
    \textbf{Details of poorly rated conversations.}
    We show here all conversations with a poor rating. Overall Satisfaction: below 3/5; Clarity, Trust and Empathy: below 2/4. Impact of Mo on negative ratings seems limited.
    }
\fontsize{8pt}{10pt}\selectfont
\renewcommand{\arraystretch}{1.5} 
\setlength{\tabcolsep}{4pt} 
\vspace{.8em}
\begin{tabular*}{\linewidth}{@{\extracolsep{\fill}}lp{5cm}p{4cm}cccc}
\toprule
Group & Description & Role of Mo & \makecell{Overall\\Satisfaction} & Clarity & Trust & Empathy \\ 
\midrule\addlinespace[2.5pt]
Control & Low rating justified. Patient asks a clear pediatric question and the physician makes a diagnosis too quickly without answering the initial question. & Not involved & 2 & 2 & 2 & 2 \\ 
Control & Low rating partly justified. Doctor is not assessing the problem because a GP is on their way to do a physical examination, and it's the best solution. GP could have been more empathetic and pedagogic. & Not involved & 2 & 3 & 3 & 2 \\ 
Control & Medium rating without apparent justification. The answer was great, and the patient seemed happy with the conversation. & Not involved & 3 & 3 & 3 & 3 \\ 
Control & Low rating justified. Patient came for psychological distress and was not redirected or provided with options. & Not involved & 3 & 2 & 2 & 1 \\ 
Control & Medium rating without apparent justification. Patient came for a complaint that needed further examination and was invited to consult in real life. & Not involved & 3 & 4 & 4 & 4 \\ 
Control & Low rating justified. Patient is concerned about their daughter. Doctors advised calling emergency services (15) without asking more questions or giving advice. & Not involved & 2 & 3 & 2 & 2 \\ 
Mo Proposed & Medium rating without apparent justification. Patient asked for pediatric advice; Mo answered well, and the doctor validated the response. & Good behavior & 4 & 3 & 2 & 3 \\ 
Mo Proposed & Low rating partly justified. Patient asked for an appointment with a specialist for a chronic issue (3 years). They requested the phone number of our doctors but were redirected to teleconsultation. & Mo promised help to find a specialist but failed to give useful advice, which might have annoyed the member. & 1 & 1 & 1 & 1 \\ 
\bottomrule
\end{tabular*}
\label{tab:conversation_details}
\end{table}

\setcounter{figure}{0}
\renewcommand{\thefigure}{S\arabic{figure}}

\begin{figure}[H]
    \centering    
    \begin{minipage}{0.24\textwidth}
        \centering
        \includegraphics[width=\textwidth]{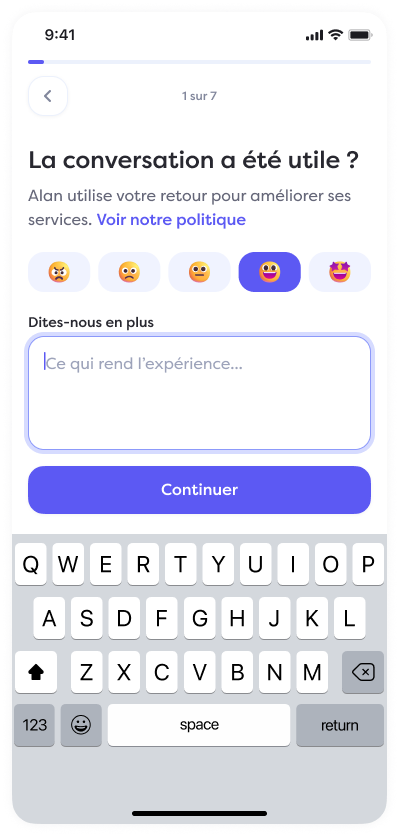}
    \end{minipage}
    \hfill
    \begin{minipage}{0.24\textwidth}
        \centering
        \includegraphics[width=\textwidth]{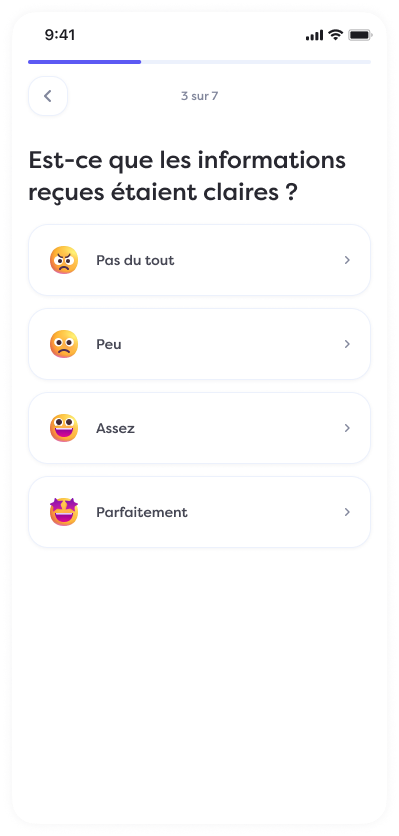}
    \end{minipage}
    \hfill
    \begin{minipage}{0.24\textwidth}
        \includegraphics[width=\textwidth]{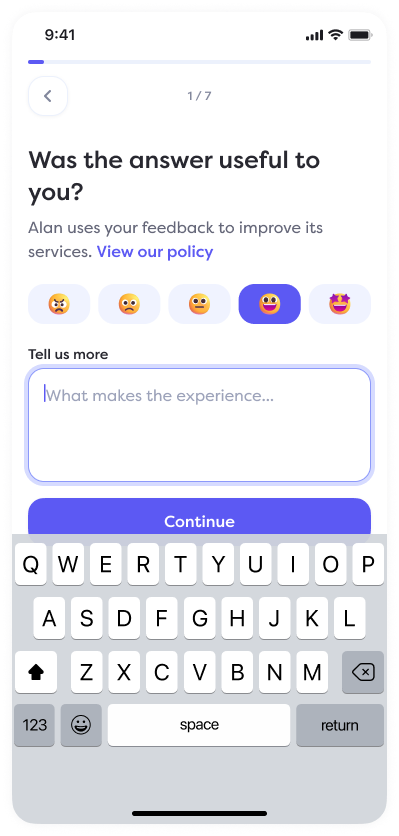}
        \centering
    \end{minipage}
    \hfill
    \begin{minipage}{0.24\textwidth}
        \centering
        \includegraphics[width=\textwidth]{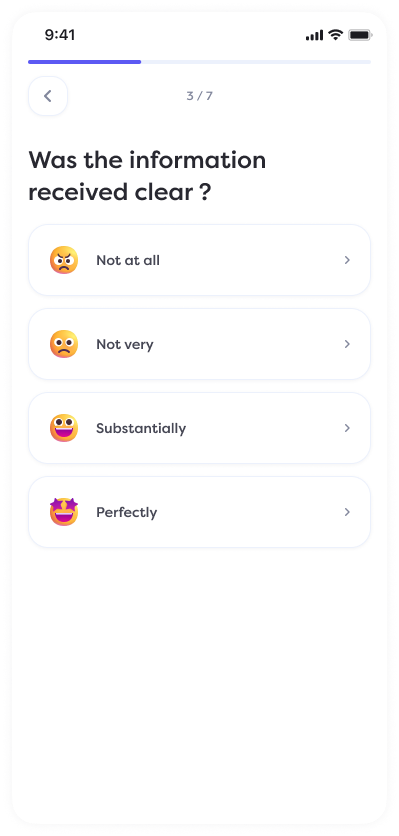}
    \end{minipage}
    \caption{Example screens for member feedback in French (orignal, left) and English (translated, right)}
    \label{fig:user_screens}
\end{figure}

\begin{figure}[H]
    \centering    
    \begin{minipage}{0.24\textwidth}
        \centering
        \includegraphics[width=\textwidth]{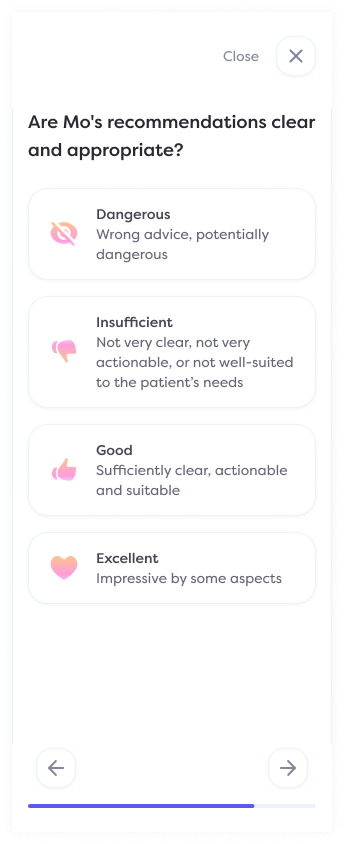}
    \end{minipage}
    \hfill
    \begin{minipage}{0.24\textwidth}
        \centering
        \includegraphics[width=\textwidth]{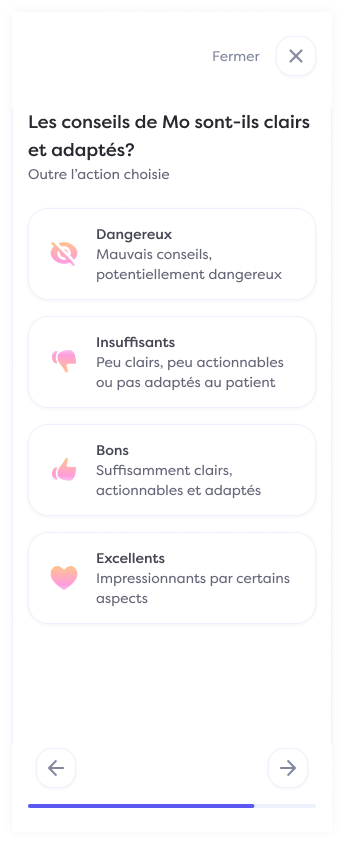}
    \end{minipage}
    \hfill
    \begin{minipage}{0.24\textwidth}
        \centering
        \includegraphics[width=\textwidth]{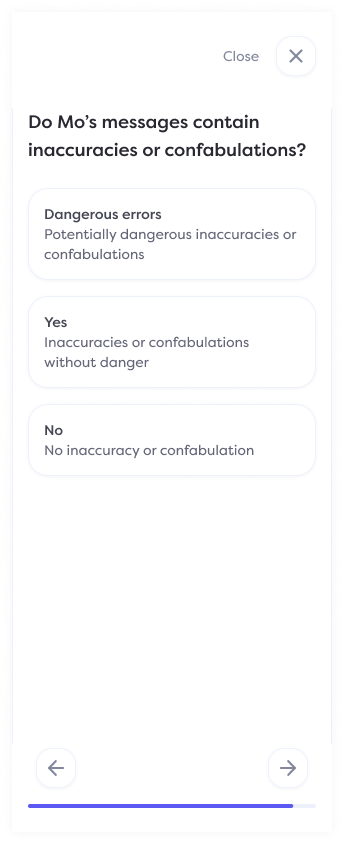}
    \end{minipage}
    \hfill
    \begin{minipage}{0.24\textwidth}
        \centering
        \includegraphics[width=\textwidth]{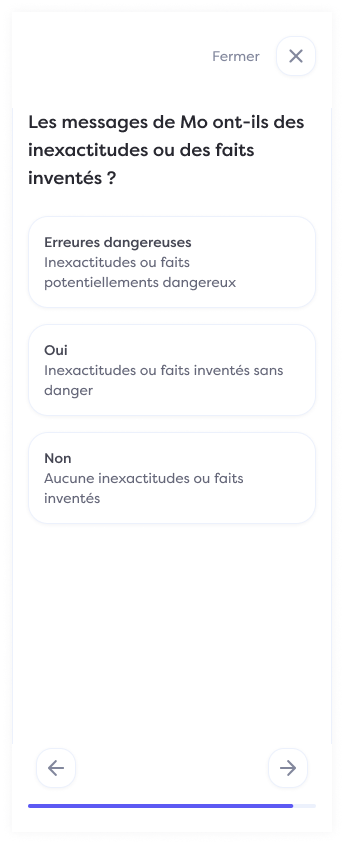}
    \end{minipage}
    \caption{Example screens for physician evaluation in French (orignal, left) and English (translated, right)}
    \label{fig:gp_screens}
\end{figure}

\end{document}